%% file: sn-article.tex
\documentclass[sn-mathphys,Numbered]{sn-jnl}

\usepackage{graphicx}%
\usepackage{multirow}%
\usepackage{amsmath,amssymb,amsfonts}%
\usepackage{amsthm}%
\usepackage{mathrsfs}%
\usepackage[title]{appendix}%
\usepackage{xcolor}%
\usepackage{textcomp}%
\usepackage{manyfoot}%
\usepackage{booktabs}%
\usepackage{algorithm}%
\usepackage{algorithmicx}%
\usepackage{algpseudocode}%
\usepackage{listings}%

\usepackage{array}
\usepackage{graphicx}
\usepackage{subcaption}
\usepackage{hyperref}
\usepackage{enumitem}
\usepackage{float} 
\usepackage{placeins} 

\theoremstyle{thmstyleone}%
\theoremstyle{thmstyletwo}%

\theoremstyle{thmstylethree}%

\begin{document}

\title[Article Title]{Phoenix: A Federated Generative Diffusion Model}








\author[1]{\fnm{Fiona Victoria} \sur{Stanley Jothiraj}}\email{fiona123@uw.edu}
\author[]{\fnm{Afra} \sur{Mashhadi}}\email{mashhadi@uw.edu}

\affil[]{\orgdiv{University of Washington Bothell}, \orgaddress{\state{Washington}, \country{USA}}}

\abstract{Generative AI has made impressive strides in enabling  users to create diverse and realistic visual content such as images, videos, and audio. However, training generative models on large centralized datasets can pose challenges in terms of data privacy, security, and accessibility. Federated learning (FL) is an approach that uses decentralized techniques to collaboratively train a shared deep learning model while retaining the training data on individual edge devices to preserve data privacy. This paper proposes a novel method for training a Denoising Diffusion Probabilistic Model (DDPM) across multiple data sources, using FL techniques. Diffusion models, a newly emerging generative model, show promising results in achieving superior quality images than Generative Adversarial Networks (GANs). Our proposed method \textit{Phoenix} is an unconditional diffusion model that leverages strategies to improve the data diversity of generated samples even when trained on data with statistical heterogeneity or Non-IID (Non-Independent and Identically Distributed) data. We demonstrate how our approach outperforms the default diffusion model in a FL setting. These results are indicative that high-quality samples can be generated by maintaining data diversity, preserving privacy, and reducing communication between data sources, offering exciting new possibilities in the field of generative AI.}

\keywords{Federated learning, Generative AI, Diffusion Models, Heterogeneous data}

\maketitle

\vspace{-0.4cm}
\section{Introduction}
    In recent years,  Generative AI   has emerged as a rapidly growing field of research 
    and development of innovative tools such as  Stable diffusion \cite{rombach2022high}, Imagen \cite{saharia2022photorealistic}, ChatGPT that are used beyond the computer science community. Generative AI refers to generating high-quality synthetic content using advanced AI techniques such as  natural language processing (NLP), computer vision (CV), and other fields of machine learning (ML). Conventional AI models were created to perform classical, regression, or clustering tasks that helped analyze existing data. What sets generative AI is its distinctive ability to generate novel content based on the patterns and information learned from the training data.

    Deep generative models (DGMs), a deep learning approach in Generative AI are a special class of probabilistic neural networks that can estimate the likelihood of each image or observation to generate new samples based on the underlying distribution. Commonly used DGM techniques include generative adversarial networks (GANs), Variational Autoencoders (VAEs), and Diffusion Models. This method of unsupervised learning has potential applications, like data augmentation, code generation, text, video and audio synthesis, anomaly detection, etc \cite{bondtaylor2022}.
    
    
     At the same time, pressing  data privacy concerns have prompted the development of a distribution training approach that differs from centralized training. Federated Learning (FL)~\cite{mcmahan2017communication} aimed to create a decentralized collaborative learning scenario wherein the user data is trained locally without the need for data to be transferred to a central server. This privacy-by-design training technique meant that several edge devices could be trained with rich representations of global data without any direct data interaction. The process involves training local data on each client and transmitting the learned parameters (model weights) from each round of training to the global model for aggregation. Hence, FL offers the potential to create a high-performing global model without the need to collect training data in a central location. However, research has shown that existing GAN models suffer from heterogeneity challenges of FL paradigm \cite{zhang2021training} \cite{ganprob}. 
     
    This paper presents a novel method to train a generative unconditional \textit{diffusion} model in a distributed way for creating  synthetic images. Specifically, we aim to address the problem of statistical heterogeneity in FL when local data is heterogeneous (referred to as non-IID). We first demonstrate that generated samples' quality and mode coverage for state-of-the-art GAN models suffer from heterogeneity in FL. Next, we propose two strategies to address statistical challenges when designing a diffusion model in FL, namely \textit{Data sharing Strategy} and \textit{Personalization \& Threshold Filtering}. The effectiveness of these strategies is evaluated through experiments to show the improvement in data quality and mode coverage over GANs. We particularly utilize unconditional diffusion models that generate data samples without specific conditioning or external factors such as prompts. These models learn the underlying probability distribution of the training data and generate new samples (based on seed value) that resemble the training data. To the best of our knowledge, no prior work has  been done on applying generative diffusion models in FL. 
    
    The algorithm name ``\textit{Phoenix}" is aptly chosen for our approach due to its transformative nature. Just as how the mythical bird rises from ashes, the diffusion model can be symbolized by its transition from chaos (pure noise) to creation (generated data). Our study provides the first step in providing a promising solution to train diffusion models through FL for image synthesis tackling the statistical heterogeneity of clients.
    
    This article is outlined as follows: Section 2 presents related works on generative diffusion models and FL. Section 3 outlines the proposed FL methodology for handling IID and Non-IID data, along with quantitative metrics. Section 4 covers implementation, experiments, and results. Finally, Section 5 discusses limitations, implications, and future directions.

\section{Related Work} 

    \subsection{Generative  Models}

    Generative Artificial Intelligence is a new class of AI algorithms that aim to generate new content or output based on the vast amount of data that it has been trained on. Researchers have highly sought them due to their exceptionally high-quality data samples generated by the models. This new family of models, also known as deep generative models (DGMs), combines the power of generative models and deep neural networks. One of the first generative models was proposed in 2014 \cite{goodfellow2014generative} called the adversarial nets framework. The framework consists of a generative model that acts against an adversary discriminative model, which learns to distinguish whether a data sample is indeed from the original data distribution. It was analogous to a counterfeit team who produces counterfeit currency notes and tries to get away with it (i.e. generative model) and the police who detect the counterfeit currency notes (i.e., discriminative model). Both the generator and discriminator are deep neural networks, and the output of the generator is fed as input to the discriminator. During backpropagation, the weights of the generator are updated through the discriminator's classification task.
  
    The paper \cite{pmlr-v37-sohl-dickstein15} originally introduced the idea of diffusion probabilistic models that aim to learn how to reverse a gradual multi-step noising or decaying process. The diffusion probabilistic model or diffusion model is a parameterized Markov chain that produces samples after finite samples. Transitions of this chain are taught to learn how to reverse a diffusion process when the chain gradually adds noise to the data until all information is destroyed. 
    The DDPM Model \cite{ho2020denoising} shows that extremely high-quality samples could be generated on the CIFAR-10 and LSUN datasets. Their work solely relied on predicting the noise found in the image, not the mean, and fixing the variance of the Gaussian noise in the forward pass. Further improvements on the DDPM model \cite{nichol2021improved} showed the importance of learning variances of the reverse diffusion process that allows for sampling with fewer forward passes without affecting sample quality.

 \subsection{Federated Learning}
 Federated learning (FL) \cite{mcmahan2017communication,lo2020litreview} is an emerging Machine learning technique that enables ML models to be trained in distributed data on edge devices without being stored on a central server.  In FL, each client retains its local training dataset, ensuring that data is not transmitted to a central server. Instead, the clients (or individual devices) are trained on the neural network models, and only the computed model updates are communicated. The FL technique can thus significantly reduce the privacy and security risks in comparison to the traditional centralized technique. As per the White House Report \cite{whitehouse2013}, the FL technique best makes use of principles such as data minimization and security. While FL offers privacy and security benefits over traditional methods, it also faces certain challenges. The training data present in clients are typically personalized and hence not representative of the global data distribution. This poses a problem of non-IID data, which is described in detail in section 2.2. Since the clients act as edge devices, there is typically limited communication, or they are frequently offline; hence, some devices train faster than others. 


    

  After each round of the local model, the computed model weights are sent to the central server for aggregation. Various model aggregation techniques are used depending on the nature of the local data and the system behavior such as FedSGD, FedAvg \cite{mcmahan2017communication}, FedProx \cite{li2020federated}, QFedAvg \cite{li2019fair} and FedOptim \cite{reddi2020adaptive}. Out of which FedAvg is mostly used as a benchmark.

\noindent{\textbf{Heterogeneity Challenges in FL:}}\label{sec:iidvsnoniid}  It is common in a FL network to have clients with different data distributions. For instance, each patient's medical history can be treated as a distinct dataset in a healthcare dataset. The data within these datasets is likely to have been drawn from different distributions. Such forms of data are called Non-IID data or Non-Independent and identically distributed \cite{wireless2020}. Although IID data is the simplest form of data distribution to deal with, it is unrealistic to assume that the data present in edge devices is IID since the clients are unique. 

Several efforts are being carried out to deal with this highly skewed non-IID data. Most methods involve data-based, algorithmic, and system-based approaches \cite{zhu2021federated}. The data-based approach preprocesses client data before training, such as using data sharing or data augmentation techniques \cite{zhao2018federated}. Algorithmic-based methods employ learning techniques like knowledge distillation, multi-task learning, and personalization layers. System-based approaches use client-based clustering methods to better understand heterogeneous data environments.


\noindent{\textbf{Stability Challenges in FL:}} In recent years, much research has been made to use GANs in an FL setting to personalize the client tasks and heterogeneous data distribution problems ~\cite{li2022federated,cao2022perfed,9694141,fan2020federated}. However, GANs have limitations such as mode collapse, convergence issues, and scalability challenges. To address these drawbacks, Diffusion Models have emerged as a more suitable approach for generative AI tasks, effectively overcoming these challenges.

\section{Methodology}

    \subsection{Diffusion Model} 
    A denoising diffusion model is a generative model that transforms noisy data, typically originating from a Gaussian distribution, into high-quality samples of the desired data distribution. A neural network backed by a U-Net model architecture \cite{ronneberger2015u} is typically used to achieve this. This model learns to gradually denoise the data, resulting in a proper image. 

    \begin{figure}[h]
      \centering
      \includegraphics[width=0.75\textwidth]{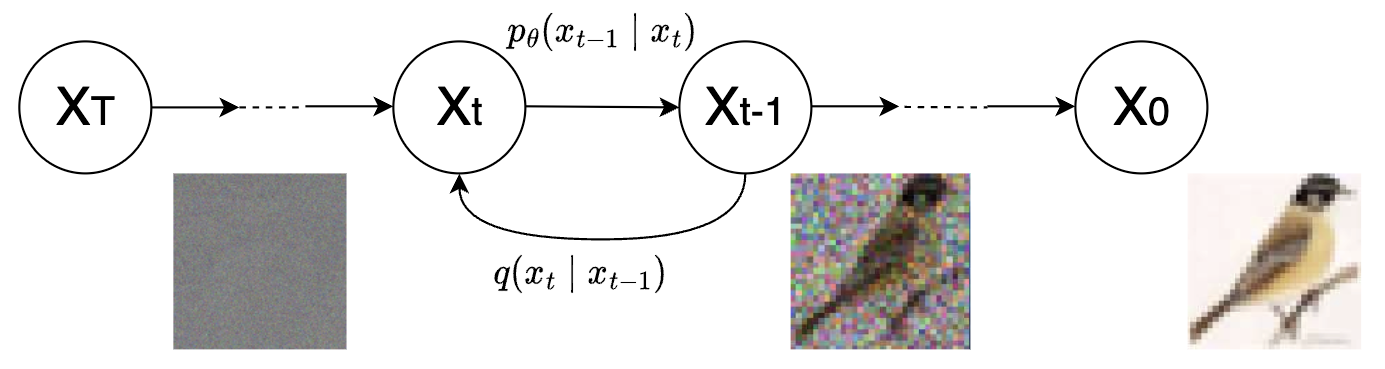}
      \caption{Directed Graphical Model of Diffusion Model \cite{ho2020denoising}}
      \label{fig:dm}
    \end{figure}
    
    The two-step process involved in diffusion models is illustrated in Figure \ref{fig:dm} and is explained as follows:
    
       \noindent \textbf{Forward diffusion process}: Given an image x0, the forward process aims to add Gaussian noise at each time step t (pre-defined) till an isotropic Gaussian distribution is achieved. As the step size t increases, the image loses all its distinguishable features, eventually becoming complete noise. 
        
        If $q(x_0)$ is the real data distribution of real images and a sample image from this distribution is $x_0 \sim q(x_0)$, then the forward diffusion process can be denoted as $q(x_t \mid x_{t-1})$. Gaussian noise is added at every time step t based on a variance scheduler $\beta$,  $0 < \beta_1 < \beta_2 < \beta_3 < .... < \beta_T < 1$ as
        \begin{equation}
        q(x_t \mid x_{t-1}) = N(x_t; \sqrt{1 - \beta_t} x_{t-1}, \beta_t I)
        \end{equation}
        The term $\sqrt{1 - \beta_t} x_{t-1}$ is the mean of the Gaussian distribution denoted by $\mu_t$ and $\beta_t$ is the variance of the Gaussian distribution denoted by $\sigma_t^2$ that is  based on sampling $\epsilon \sim N(0, I)$. Hence the next image can be calculated by
        \begin{equation}
        x_t = \sqrt{1 - \beta_t} x_{t-1} + \sqrt{\beta_t} \epsilon
        \end{equation}
        The forward diffusion process starts with an image $x_0$ and adds noise in T steps $x_1, ..., x_t, ...,x_T$ until only pure Gaussian noise remains.
    
         \noindent \textbf{Reverse diffusion process}: A neural network (typically U-Net) is trained to denoise the images containing pure noise till the actual image is retrieved. 


        Figure \ref{fig:dm} illustrates that knowing the conditional distribution $p(x_{t-1} \mid x_t)$ allows effortless image reconstruction. However, since this distribution requires knowledge of all training images, we employ Neural Networks to learn the approximate conditional probability distribution $p_\theta(x_{t-1} \mid x_t)$, with $\theta$ representing the learned parameters. The neural network learns the mean and variance of these distributions for efficient image recovery and reconstruction.
        

    \subsubsection{Hyperparameters}
    \begin{itemize} 
    \item \textbf{Variance Scheduler}: The rate at which noise is added to the images is controlled by the \textbf{variance scheduler $\beta$}. This pre-defined parameter can be a constant integer or chosen as a scheduler over T timesteps. Earlier works \cite{ho2020denoising} employ a linear schedule from $\beta_1 = 10^{-4}$ to $\beta_T = 0.02$. However, this linear schedule's effectiveness is limited for smaller resolutions like 32 x 32 and 64 x 64, where it introduces excessive noise without improving sample quality. The improved work \cite{nichol2021improved} introduced a cosine scheduler with linear drop-off in the middle of the decay process, while preserving information at the start and end, leading to slower destruction of useful information and improved accuracy in prediction.
    

    \item \textbf{Diffusion Steps}:
    The number of diffusion steps, denoted as T, determines the decay process. Prior works \cite{ho2020denoising} used T = 1000, while the improved DDPM \cite{nichol2021improved} trained models with 4000 steps. To address the time-consuming nature, they proposed sampling intermediate steps of 25, 50, 100, 1000, etc.

    \item \textbf{Learning Rate}:
    The learning rate (lr) regulates the weight update speed during training. It requires careful selection to avoid sub-optimal solutions (with a high learning rate) or getting stuck at local minima (with a low learning rate) during gradient descent.
    \end{itemize} 

    \FloatBarrier
    
  \subsection{\textit{Phoenix} Algorithm} 

  We propose \textit{Phoenix}, a novel diffusion neural network model that is specifically designed to be trained in a FL setting. The primary focus of our work is to explore the implications and effects of utilizing such a model on two distinct data scenarios: independent and identically distributed (IID) data and non-IID data. The default Phoenix framework utilizes 10 clients, each with 100 local epochs and 10 server rounds. The training process involves splitting the CIFAR-10 dataset into train and test sets, further dividing the training data among the clients. Clients train locally using diffusion models with a U-Net backbone architecture. The model weights are averaged in the global server after 100 local epochs. Gradient descent is employed for weight updates, with generated image samples saved locally for evaluation. This iterative process ensures continuous improvement and the generation of high-quality synthetic data.

    \subsubsection{Data Sharing Strategy}
    One way to solve the diversity issues faced by Non-IID data is to utilize a data-sharing strategy that uses a small subset of global data shared between all devices before FL training. By sharing a small subset of global data, the overlap of data distribution across clients can be increased, thereby improving the overall model performance. Prior experiments \cite{zhao2018federated} show that when using a CIFAR-10 dataset trained on CNN models, they achieved a 30\% increase in test accuracy when using 5\% of the globally shared data. 

    The stepwise process of data sharing strategy for a CIFAR-10 dataset is explained as follows,
    \begin{enumerate}
        \item The CIFAR-10 training set, comprising 50,000 image samples, is divided into two parts: the Client (C) with 40,000 examples and the Server (S) with 10,000 examples.
        \item The Client (C) part is split into ten clients chosen as the number of clients that participate in the FL training process. The partition is usually a case of 1-class or 2-class non-IID Data. ie. The number of classes in a client is one or utmost two, which helps simulate the scenario of non-IID Data.
        \item The Server (S) part is split into ten groups as a globally shared data (G) based on a factor $\beta$ that is quantified by $\beta = \frac{\left\lvert G \right\rvert}{\left\lvert C \right\rvert} \times 100\%$. $\beta$ can range from 2.5\% to 25\%.
        \item Next, the global warmup model is trained on globally shared data (G) for the required number of epochs till it reaches the desired accuracy.
        \item From the globally shared data (G), a fraction of data known as $\alpha$ is merged with every client part (C) for FL training. $\alpha$ of zero means that no data from G is shared with client, and $\alpha$ of 100 means that all the data from G is merged with every client.
    \end{enumerate}
    


    
    \subsubsection{Personalization layers \& Threshold Filtering}
    \textbf{Personalization layers}:
    Statistical Heterogeneity \cite{arivazhagan2019federated} i.e., data distribution is different for different clients and typically impacts the performance of ML models. The personalization layer-based approach \cite{metapersonalized} is one popular way of handling different clients with very different data distributions. It aids in the customization of models to individual users or devices without compromising user privacy. To better capture the features/properties each client learns, the deep neural networks are divided into personalization and base layers. The base layers are typically co-trained by the clients and take part in the model aggregation stage, whereas the personalization layers \cite{perfed} do not leave the individual clients.

    We use U-Net \cite{ronneberger2015u}, a convolution neural network, as the backbone model architecture for diffusion models. U-Net contains a total of eight basic blocks: 4 blocks on the encoder side and 4 on the decoder side. To better understand the effects of personalization layers, we consider the last basic block on the decoder side containing convolution layers to be the personalization layers.
    
\begin{figure}[htbp]
      \centering
      \includegraphics[width=1.0\textwidth]{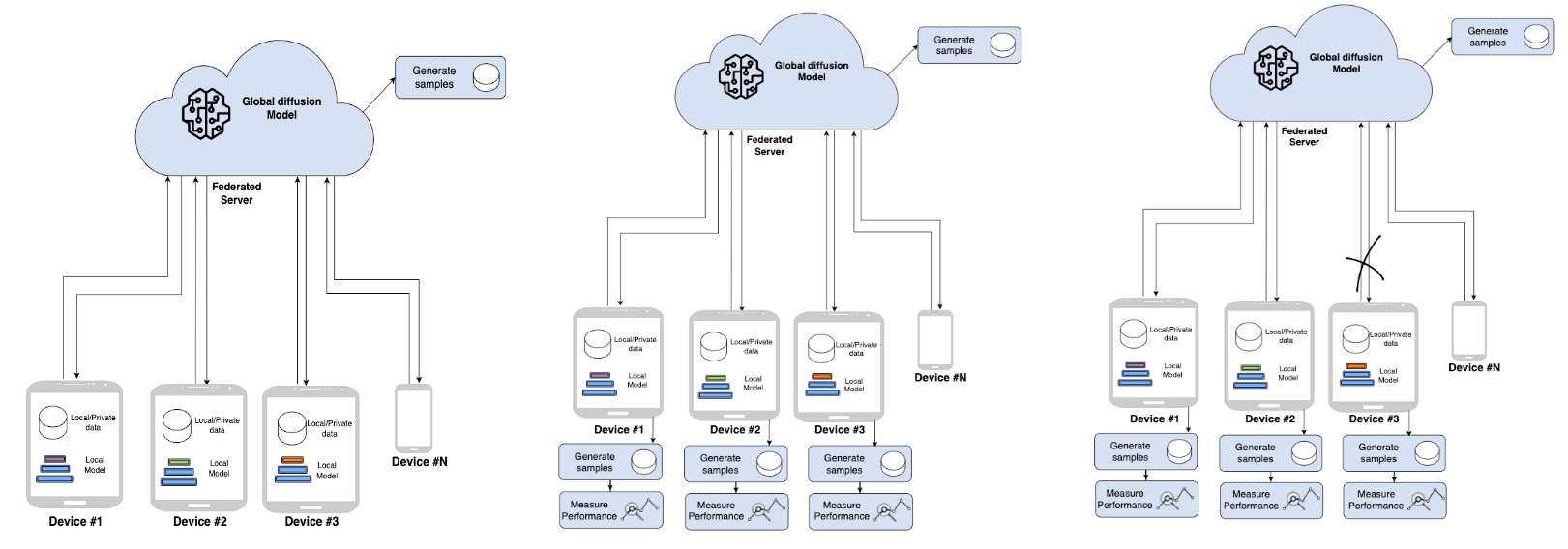}
      \caption[Illustration of Personalization layers \& Threshold Filtering strategy in FL setting]{Illustration of Personalization layers \& Threshold Filtering strategy in FL setting \textbf{Step 1}: Personalization layers, 
      \textbf{Step 2}: Generate samples to monitor performance and \textbf{Step 3}: Disconnect underperforming clients}
      \label{fig:per_thres}
    \end{figure}

    \textbf{Threshold Filtering}:
    We propose another strategy to improve the overall model performance in the global model by means of a threshold filtering technique. In the context of FL, threshold filtering can be used to remove (or) disconnect underperforming clients from the active training process (Figure \ref{fig:per_thres}). Precision and recall are the performance metrics to determine if a client is performing below optimal levels. Clients that do not meet the pre-set/average values of the threshold are removed either temporarily or permanently. This can be achieved in two ways: excluding from a particular server communication round or removing from the training altogether. 
    
    In our method of threshold filtering, we use a warning mechanism to disconnect clients only if they underperform consistently for more than two local trainings. By implementing this two-strike approach of threshold filtering, the FL system can help improve performance, efficiency, and fairness while still maintaining its privacy and scalability advantages. This method of threshold filtering technique offers several other benefits as listed below,
    \begin{itemize}[itemsep=0pt]
        \item Robustness: We give all clients a second chance and do not penalize clients immediately for temporary performance drops due to factors such as fluctuations in data quality. 
        \item Fairness: By allowing underperforming clients to participate in further training rounds before being removed, there is a fair evaluation of performance. This eliminates  premature exclusion  that could arise due to connection/communication/data issues.
        \item Dynamic Adaption: The constant monitoring of a FL system helps keep track of clients' performance even when the client population or data distribution changes.
        \item Improved Global Model Performance: By removing underperforming clients, the global model performance may be less impacted by noise or biased data.
        \item Faster Convergence: Underperforming clients in a training process require more server communication rounds to achieve optimal levels of performance. By removing such clients, the models are trained on high-quality, relevant data, thereby training efficient and faster convergence of the model.
        \item Optimized Resource Utilization: The time and resources spent in training  clients that perform at suboptimal levels can be eliminated, which is a significant overhead in FL.
    \end{itemize}
    


 \subsection{Dataset and Data Preparation}
  The generative diffusion models were trained on the CIFAR-10 dataset \cite{cifar10}. The CIFAR-10 dataset consists of 60k samples in 10 different classes of size 32 x 32, split into 50k training images and 10k test images. The \textit{Phoenix} framework trains the diffusion models using the 50k samples and evaluates the performance of these models using the 10k test images. CIFAR-10 is a commonly chosen benchmark dataset for evaluating neural network models in various domains.
 
    \paragraph{IID Data} The Independent and Identically Distributed (IID) data is simulated during training. The CIFAR-10 dataset is distributed among the k-clients as the diversity of image samples in each client is almost equal. Although the occurrence of IID data distribution in real-world applications is not practical, this is done to understand the behavior of diffusion model systems.
    


    \paragraph{Non-IID Data}
    There exist three different ways to make data behave non-identically distributed across clients \cite{li2022federatedstudy} \cite{Li2020LotteryFLPA}: feature distribution skew, label distribution skew and quantity skew. The experiments in \textit{Phoenix} use label distribution skew that is 
    given a specific label, each client owns highly different amounts of samples corresponding to that label.  We follow the distribution suggested by  \cite{zhao2018federated} where there are 2 class Non-IID data. A 2-class non-IID case ensures that each client contains samples belonging to utmost two classes.

    
    
    
    

  \subsection{Metrics}
       \textbf{Inception score} - The IS score uses the Inception V3 Network pre-trained on the ImageNet dataset to calculate the single floating point number score to measure the diversity and fidelity of images.
        
  \noindent      \textbf{Fréchet Inception Distance score} - FID \cite{heusel2018gans} measures the Gaussian distribution between real and generated images. 
        
   \noindent       \textbf{Precision and Recall} - Precision denotes what fraction of the generated samples is realistic, and recall measures what fraction of the training set manifold the generator covers. Both these metrics \cite{sajjadi2018assessing} \cite{kynkaanniemi2019improved} can effectively measure the fidelity and diversity of the samples.

 \noindent         \textbf{Classification Distribution} - We propose using a pre-trained state-of-the-art classification model \cite{lanet} LaNet that quotes a top-1 accuracy of 99.03\% on the CIFAR-10 dataset using the NASNet search space. By comparing the classification results of the generated samples with those from the test set, we can effectively measure the diversity of generated samples. If the generated samples are diverse and realistic, we would expect to see a similar distribution of classes between the two sets (generated and test set). However, if the generated samples are not as diverse as expected, we would expect to see a skewed distribution of classes. Using a pre-trained classification model is an objective measure to evaluate generative models' performance.

\section{Results}
    \subsection{Baseline Using GANs}
     
     We compare our proposed methods of diffusion model trained in a FL setting with that of GANs.  For the purpose of comparison, we chose the GAN architecture - Deep convolutional generative adversarial networks (DCGAN) \cite{radford2016unsupervised}. DCGAN is a good baseline since it is a popular choice of GAN architectures for image generation tasks. It also addresses several limitations of the original GAN architecture improving data quality and stability during training.
    

    We experiment with three different training modes, centralized machine learning \cite{ksuryateja} and FL with IID data and FL with Non-IID Data, using DCGAN as our backbone architecture. We use the CIFAR-10 dataset containing 50,000 training data for these experiments. Pre-processing was performed on all training images to resize, normalize and convert the images to tensor. All models were trained using a mini-batch size of 128 and optimized using the Adam optimizer. We set the learning rate as 0.0002 and the momentum term $\beta$ as 0.5 to stabilize training. For FL, we used a 2 class Non-IID simulated data \cite{jeremy-noniid} and set the following parameters prior to training: Server Rounds = 10, Aggregation Strategy = FedAvg, Training Set = 50k samples, Optimizer = Adam, lr=0.0002, Epochs = 50. The results of the training are tabulated in Table \ref{tab:dcgan}.

    Additionally, in order to visualize the coverage of classes from synthesized data, we perform a classification task on the 10,000 generated samples. An image classification task using the \textbf{LaNet} model was performed on our generated sample since this model was also trained on the CIFAR-10 dataset. This open-source project from Facebook Research has been ranked \#17 in image classification on CIFAR-10 with an accuracy of 99.03, containing over 44.1M parameters. We compare this against the test set of the CIFAR-10 dataset containing 10,000 samples with approximately 1000 samples per class, such as airplanes, cars, birds, cats, deer, dogs, frogs, horses, ships, and trucks. Our assumption is that the LaNet model's classification task on these 10,000 samples would also result in the same result - 1000 samples are predicted per class. This result serves as our baseline, and we aim to match the same distribution of synthesized images with our proposed models. 

\begin{figure}[!htbp]
        \centering
        \includegraphics[width=0.40\linewidth]{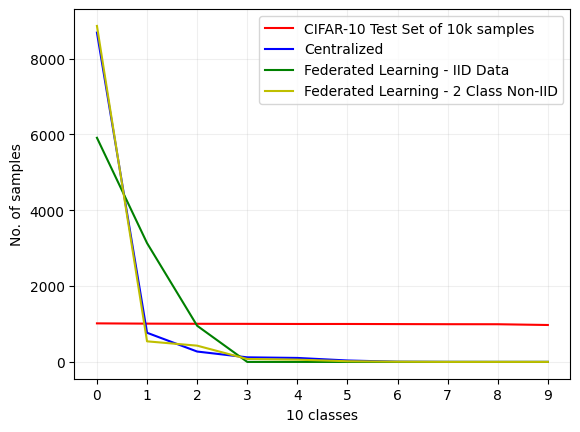}
        \hspace{0.02\linewidth} 
        \includegraphics[width=0.40\linewidth]{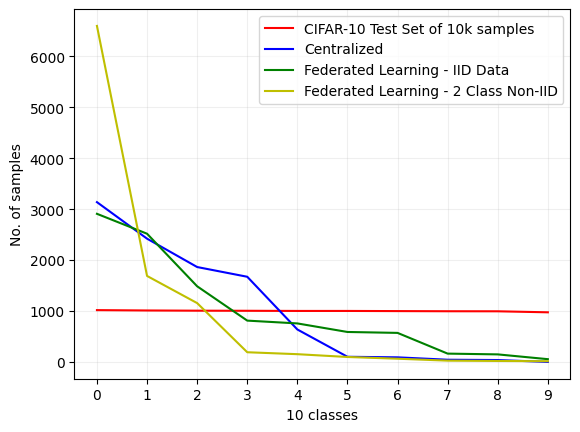}
        \caption{Distribution of generated sample classes based on classification results from
            LaNet Model sorted in descending order for DCGAN (left) and default diffusion model(right)}
        \label{fig:sample_distributions_gan_dm}
    \end{figure}

    Based on our experiment results in Table \ref{tab:dcgan}, it appears that the centralized mode of training outperforms the FL technique in terms of metrics like FID, IS, Precision, and Recall. Furthermore, it seems that the performance of FL methods greatly depends on the type of data used. ie. The performance of FL on non-IID is worse than that with IID data since it is influenced by the uneven distribution of data across the participating devices. This can be visualized by the distribution of generated data (10k samples) for different modes as shown in Figure \ref{fig:sample_distributions_gan_dm}.
    
\begin{figure}[htbp]
  \centering
  \begin{minipage}[b]{0.23\textwidth}
    \centering
    \includegraphics[width=\textwidth]{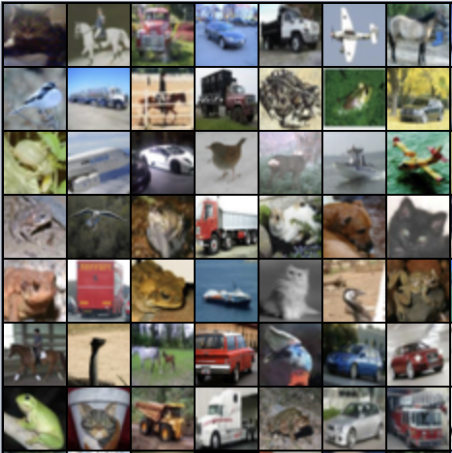}
    \caption*{Original Data from CIFAR-10}
    \label{fig:figure1}
  \end{minipage}
  \hfill
  \begin{minipage}[b]{0.23\textwidth}
    \centering
    \includegraphics[width=\textwidth]{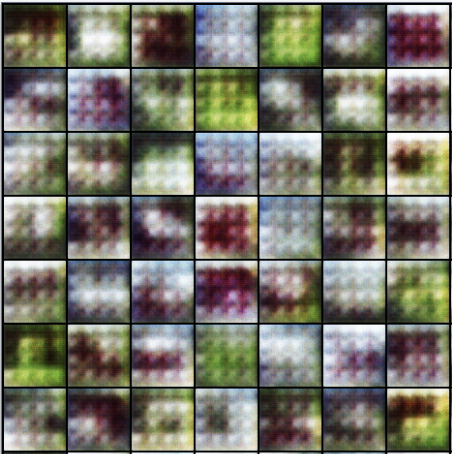}
    \caption*{Generated samples at Step 500}
    \label{fig:figure2}
  \end{minipage}
  \hfill
  \begin{minipage}[b]{0.23\textwidth}
    \centering
    \includegraphics[width=\textwidth]{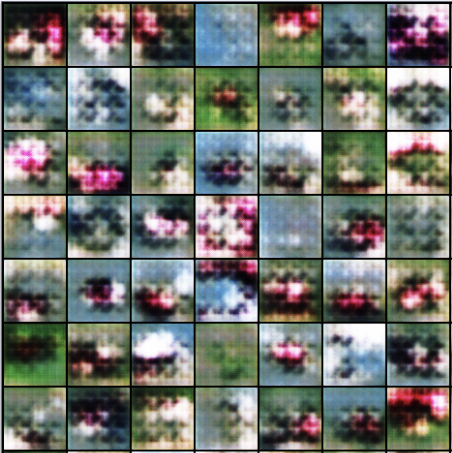}
    \caption*{Generated samples at Step 1000}
    \label{fig:figure3}
  \end{minipage}
  \hfill
  \begin{minipage}[b]{0.23\textwidth}
    \centering
    \includegraphics[width=\textwidth]{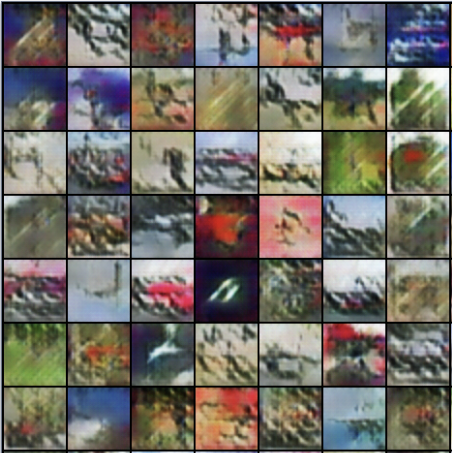}
    \caption*{Generated samples at Step 5000}
    \label{fig:figure4}
  \end{minipage}
  
  \vspace{0.5cm}
  
  \begin{minipage}[b]{0.23\textwidth}
    \centering
    \includegraphics[width=\textwidth]{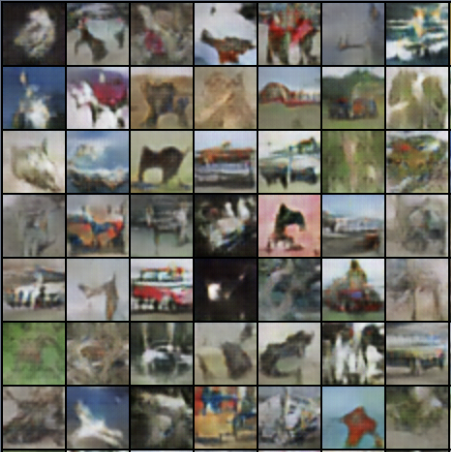}
    \caption*{Generated samples at Step 10000}
    \label{fig:figure5}
  \end{minipage}
  \hfill
  \begin{minipage}[b]{0.23\textwidth}
    \centering
    \includegraphics[width=\textwidth]{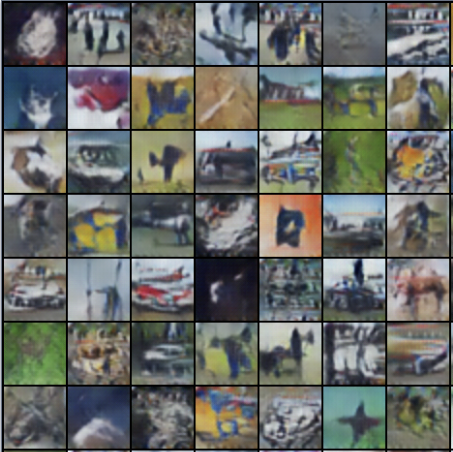}
    \caption*{Generated samples at Step 15000}
    \label{fig:figure6}
  \end{minipage}
  \hfill
  \begin{minipage}[b]{0.23\textwidth}
    \centering
    \includegraphics[width=\textwidth]{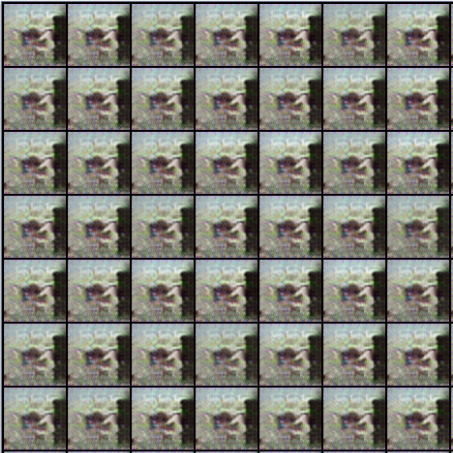}
    \caption*{Generated samples at Step 25000}
    \label{fig:figure7}
  \end{minipage}
  \hfill
  \begin{minipage}[b]{0.23\textwidth}
    \centering
    \includegraphics[width=\textwidth]{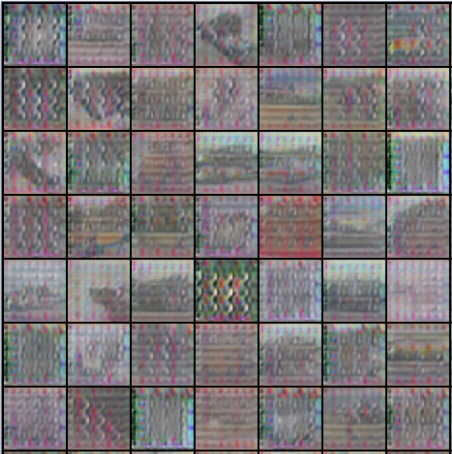}
    \caption*{Generated samples at Step 40000}
    \label{fig:figure8}
  \end{minipage}

  \caption[Samples generated during DCGAN training]{Samples when training the DCGAN from step 500 to 40000. Each image (except the first one, which is the original test data set from CIFAR-10)}

  \label{fig:dcgantrain}
 \end{figure}

    The experiments were conducted in the \textbf{NVIDIA Tesla K80 Accelerator} with a memory size of 12 GB which is optimal for DCGAN training. The overall time to train the centralized machine learning is approximately 5 hours, whereas the FL takes approximately 2 hours per server round of training. DCGANs contain 3.5 million parameters in the generator model and 2.7 million parameters in the discriminator model, totaling to approximately \textbf{6.2 million parameters}. This makes GAN models much faster to train and generate samples when compared to Diffusion Models. However, GAN contains a non-convex optimization problem, meaning that there are several local optima that the optimization algorithm can get stuck in. This means that the generator and discriminator models are unable to converge to a stable solution, and hence the performance can oscillate or drop off completely (Figure \ref{fig:dcgantrain}).

    \begin{table}[h]
        \caption{DCGAN training on CIFAR-10 data}\label{tab:dcgan}
        \begin{tabular}{@{}cccccc@{}}
        \toprule
        \textbf{Type of Training} & \textbf{FID} & \textbf{IS} & \textbf{Precision} & \textbf{Recall} \\
        \midrule
        Centralized & 0.5632 & 3.9822 $\pm$ 0.2800 & 0.9243 & 0.4951 \\
        Federated Learning - IID Data & 1.9219 & 3.4588 $\pm$ 0.2313 & 0.8189 & 0.0744 \\
        Federated Learning - 2 Class Non-IID & 6.9550 & 1.5390 $\pm$ 0.0388 & 0.1772 & 0.001 \\
        \botrule
        \end{tabular}
    \end{table}


    \subsection {\textit{Phoenix} Algorithm}
    We conducted experiments by setting up the state-of-the-art Denoising Diffusion Probabilistic Models (DDPM) \cite{ho2020denoising} \cite{nichol2021improved} in a distributed manner. We use the hyperparameters and architecture topology as mentioned in their work,

    \begin{itemize}[itemsep=0pt]
        \item Four ResNet blocks are used for downsampling
        \item Four ResNet blocks are used for upsampling
        \item Cosine schedule and Adam optimizer with Mean Squared Error
        \item The learning rate is set to $1\times 10^{-4}$
        \item Performance metrics are computed using 10K samples, as in prior works.
    \end{itemize}

For implementation we use Flower framework \cite{beutel2020flower}, along with PyTorch Library, DCGAN Library \cite{ksuryateja}, Diffusers Library \cite{von-platen-etal-2022-diffusers}. The experiments were conducted in the \textbf{GeForce RTX 2080 Ti} with a memory size of 11 GB. The overall time to train the centralized machine learning is approximately 2 hours/epoch, whereas the FL takes approximately 11 hours per server round of training. DDPM model contains nearly \textbf{35 million parameters} which is six times the size of a DCGAN model. This makes diffusion models relatively slower to train and generate samples when compared to GANs. The visual sample quality of a Diffusion Model trained on the CIFAR-10 dataset is shown in Figure \ref{fig:ddpm_samples}. 
    Based on the results of the experiment in Table \ref{tab:dmtable}, it appears that the centralized mode of training outperforms the federate learning technique in terms of metrics like FID, IS, Precision, and Recall. Given that the paper focuses on improving the sample quality of FL methods, we focus on those experiments over that of centralized methods. Since the performance of 2-class non-iid on FL is significantly poorer in terms of precision and recall, we propose two potential methods to improve the performance of the non-IID method when generating synthetic data.

    \begin{table}[h]
        \caption{Diffusion Model default training on CIFAR-10 dataset}\label{tab:dmtable}
        \begin{tabular}{@{}cccccc@{}}
        \toprule
        \textbf{Type of Training} & \textbf{FID} & \textbf{IS} & \textbf{Precision} & \textbf{Recall} \\
        \midrule
        Centralized & 0.7842 & 3.9429 $\pm$ 0.1620 & 0.8956 & 0.4717 \\
        Federated Learning - IID Data & 0.1106 & 6.0631 $\pm$ 0.3753 & 0.8144 & 0.6628 \\
        Federated Learning - 2 Class Non-IID & 0.6351 & 4.5087 $\pm$ 0.3275 & 0.3344 & 0.5453 \\
        \botrule
        \end{tabular}
    \end{table}

    \subsubsection{Data Sharing Strategy}
    The data-sharing strategy is one of the proposed approaches to overcome the issue of statistical heterogeneity (non-iid) in generative models using FL. Here, the participating devices share a subset of their data with each other to create a more representative and diverse dataset prior to training \cite{zhao2018federated}.
    
    We divided the CIFAR-10 training dataset (50,000 samples) into a client (C) and server (S) part, each consisting of 40,000 and 10,000 samples, respectively. Next, we split the client part into non-overlapping partitions and randomly assigned these partitions to each participating client containing two classes of images per client. The server part is split into partitions (globally shared data G), and only a portion of the partition is used for global training which is quantified by $\beta$ and can range from 2.5\% to 25\%. Once the global warmup model is trained, a fraction of G (quantified by $\alpha$) is merged with the client part (C) for FL training.

        The data-sharing strategy experiment was conducted on ten clients using the FedAvg aggregation strategy with a Cosine scheduler and 1000 decay steps. Table \ref{tab:dss_10clients} shows the results of each round for different values of $\alpha$ and $\beta$. Here $\beta$ represents the percentage of data from the server part (S) that was used for training at each client, and $\alpha$ represents the percentage of data from $\beta$ used for training. 
        
        
        It can be inferred that increasing the percentage of data shared among clients (i.e., increasing $\beta$) generally leads to lower FID (i.e.,  better image quality), and higher IS, Precision and Recall values. However, with the extreme 2-class non-IID data and a lower $\beta$ of 5\%, we can still achieve high precision and recall of 0.82 and 0.62, respectively, that outperforms the default training model with precision and recall of 0.33 and 0.54 respectively. This indicates that not much of the globally shared data is needed to improve the performance of synthesized images.
        
        Moreover, it was observed that the entire server part (C) was not necessary to distribute to all clients to achieve improved performance. Instead, a random portion quantified by $\alpha$  could achieve similar performance. Since $\beta$ = 25\% yields the best performant model, we evaluated its combinations by modifying the value of $\alpha$ ranging from 25\% to 100\%. By distributing only 25\% (i.e. additional 2.5K samples per client) of the server part (S) data, we could achieve precision and recall of 0.82 and 0.67, respectively. Additionally, by increasing the value of $\alpha$ to 100\% (additional 10K samples per client), we achieved the best model through our data-sharing strategy with a precision and recall of 0.83 and 0.70. These results reveal that data distribution of class with a $\beta$ = 25 and $\alpha$ = 100 improves by over 30\%.
        
 These results suggest that sharing a higher percentage of data among clients can improve image quality, but increasing the percentage beyond a certain threshold does not yield significant improvements. However, the optimal values of $\alpha$ and $\beta$ may vary depending on the specific use case model and dataset.


        \begin{table}[h]
            \caption{Data sharing strategy experiment results with ten clients}\label{tab:dss_10clients}
            \begin{tabular}{@{}ccccccccc@{}}
            \toprule
            \textbf{$\boldsymbol{\beta}$ (\%)} & $\boldsymbol{\alpha}$ (\%) & \textbf{Server} & \textbf{Client} & \textbf{Test Set} & \textbf{FID} & \textbf{IS} & \textbf{Precision} & \textbf{Recall} \\
            \midrule
            5 & 100 & 2K & 6K & 10K & 0.2858 & 6.0624 $\pm$ 0.1583 & 0.8206 & 0.6240 \\
            15 & 100 & 6K & 10K & 10K & 0.0653 & 6.3676 $\pm$ 0.1820 & 0.7555 & 0.6930 \\
            \midrule
            25 & 100 & 10K & 14K & 10K & 0.0763 & 6.5758 $\pm$ 0.3126 & 0.8303 & 0.7006 \\
            25 & 25 & 10K & 6.5K & 10K & 0.2920 & 5.5283 $\pm$ 0.6442 & 0.8186 & 0.6701 \\
            25 & 50 & 10K & 9K & 10K & 0.2047 & 5.9563 $\pm$ 0.2638 & 0.8278 & 0.6783 \\
            25 & 75 & 10K & 11.5K & 10K & 0.1435 & 6.0122 $\pm$ 0.4829 & 0.8282 & 0.6934 \\

            \botrule
            \end{tabular}
        \end{table}
       

        \begin{figure}[htbp]
        \centering
            \includegraphics[width=0.40\textwidth]{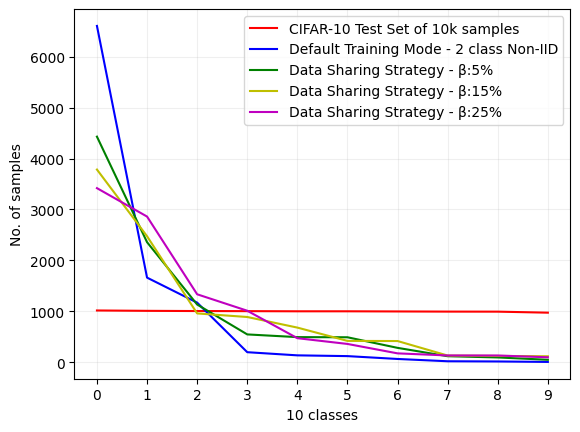}
              \includegraphics[width=0.40\textwidth]{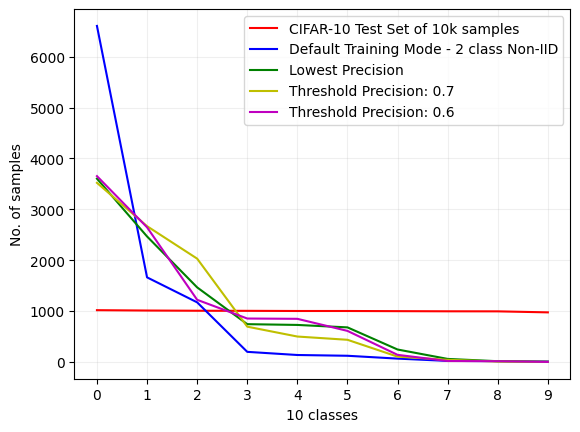}
            \caption{Distribution of generated sample classes based on classification results from LaNet Model sorted in descending order for  Data Sharing strategy (left) and  Personalization layers \& Threshold Filtering strategy (right)}
            \label{fig:dss_per_dist}
        \end{figure}
        
        
    \subsubsection{Personalization layers \& Threshold Filtering}
    The \textit{Phoenix} algorithm also leverages personalization and threshold filtering techniques to tackle statistical heterogeneity. By incorporating personalization layers during training, the FL model can better adapt to the participating devices, thereby providing personalized  generated content. We divide the diffusion model into personalization layers and base layers. Since U-Net is the chosen model architecture for the reverse diffusion process, we choose the personalization layers from this model. Our experiments designate the last basic block with convolutional layers as the personalization layers \cite{baolongnguyenmac}.

    \begin{table}[h]
        \caption{Diffusion model training with Personalization layers \& Threshold Filtering on CIFAR-10 dataset with 2 class Non-IID}\label{tab:dmper}
        \begin{tabular}{@{}ccccc@{}}
        \toprule
        \textbf{Client Drop Approach} & \textbf{FID} & \textbf{IS} & \textbf{Precision} & \textbf{Recall} \\
        \midrule
        Lowest Precision & 0.7607 & 5.3639 $\pm$ 0.3971 & 0.5920 & 0.7623 \\
        Threshold Precision: 0.7 & 1.2242 & 5.4681 $\pm$ 0.5142 & 0.5599 & 0.7614 \\
        Threshold Precision: 0.6 & 4.2548 & 2.6210 $\pm$ 0.1867 & 0.5337 & 0.259 \\
        \botrule
        \end{tabular}
    \end{table}

    Furthermore, in a scenario where the data is non-iid, it is possible that some samples provided by participating clients do not contribute value to the global server model. In the context of FL, threshold filtering is a technique to disconnect clients that exhibit subpar precision and recall during training. Since our experiments are simulation-based, new samples from clients are unlikely to be provided during training. Therefore, we permanently disconnect clients from participation in future rounds rather than temporarily disconnecting them. 
    The details of our experiments are explained as follows,
        \begin{itemize}[itemsep=0pt]

        \item Server Round 1: All clients train in default mode and save the last basic block as the personalization layer.

        \item Server Round 2 onwards: Clients retrieve the stored personalization layers before local training.

        \item Server Round 5 onwards: Clients generate 1000 samples, compute precision and recall, and log the metrics for server decisions.

        \item Warn: Clients with lower precision or below the threshold are marked separately.

        \item Monitor and Disconnect: Marked clients were closely monitored and disconnected if performance remains poor for two consecutive rounds.

        
    \end{itemize}

\begin{figure}[htbp]
        \centering
        \includegraphics[scale=0.5]{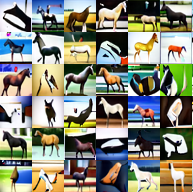}
        \hspace{0.02\linewidth}
        \includegraphics[scale=0.5]{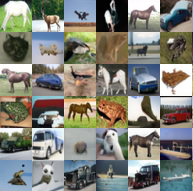}
        \hspace{0.02\linewidth}
        \includegraphics[scale=0.5]{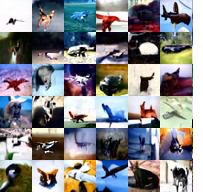}
        \caption{Generated Samples from Phoenix with Default Diffusion Training, Data Sharing Strategy, and Personalizing Strategy respectively} 
        \label{fig:ddpm_samples}
    \end{figure}

    
    The experiment was conducted with ten participating clients for ten server rounds using the federated aggregation (FedAvg) strategy with a Cosine scheduler and 1000 decay steps. Table \ref{tab:dmper} and Figure \ref{fig:dss_per_dist} show the results of Diffusion model training with Personalization layers \& Threshold Filtering on the CIFAR-10 dataset. The experiment results show that the client drop approach using the lowest precision yielded the best performance among the three approaches. This is primarily because the lowest precision approach provides all clients with equal opportunities for local training and only disconnects clients with the lowest precision in each server round. On the other hand, the other two methods employ pre-defined values, which may not be practical during training since precision and recall tend to fluctuate between server rounds. Hence, pre-defined value methods are not effective strategies for improving performance. Furthermore, the addition of personalization layers during training enhances the model's ability to adapt to the unique characteristics of each participating device. This is evident in its overall performance compared to the default training of diffusion models. The visual sample quality of a Diffusion Model trained on the CIFAR-10 dataset with Personalization layers \& Threshold Filtering Strategy is shown in Figure \ref{fig:ddpm_samples}. 
    
    An additional observation that can be made is regarding the distribution of generated samples for the different methods, as depicted in Figure \ref{fig:sample_distributions_gan_dm}, and Figure \ref{fig:dss_per_dist}. The ideal scenario for generated samples will be if they are closely aligned with CIFAR-10 Test set line consisting of 10k samples equally distributed among the 10 classes. However, the unconditional diffusion model fails to reproduce this ideal scenario.  Due to its random seed initialization for generating images, it is highly improbable for the model to effectively cover all the classes observed during training. However, we show how our proposed methods try to minimize the gap between a  biased mode coverage and the ideal flat-line scenario.

    \begin{table}[h]
        \caption{Performance comparison of generative model training methods using FL for non-IID data}\label{tab:comparison}
        \begin{tabular}{@{}ccccc@{}}
        \toprule
        \textbf{Model/Technique} & \textbf{FID} & \textbf{IS} & \textbf{Precision} & \textbf{Recall} \\
        \midrule
        Deep Convolutional GAN (DCGAN) & 6.9550 & 1.5390 & 0.1772 & 0.001 \\
        Denoising Diffusion Probabilistic Models (DDPM) & 0.6351 & 4.5087 & 0.5453 & 0.3344 \\
        Data Sharing Strategy DDPM - \textbf{Ours} & 0.0763 & 6.5758 & 0.8303 & 0.7006 \\
        Personalization \& Threshold Filtering DDPM - \textbf{Ours} & 0.7607 & 5.3639 & 0.592 & 0.7623 \\
        \botrule
        \end{tabular}
    \end{table}
    
\section{Discussion}
\subsection{Summary}

    This paper introduces a novel technique called the Phoenix for training unconditional diffusion models in a horizontal FL setting, which, to our knowledge, has not been previously explored. Its primary objective is to address the issue of mode coverage commonly encountered with non-IID datasets. 

    
    Our methods demonstrated superior model performance compared to existing techniques, such as Deep Convolutional GANs in FL and Diffusion Models in Centralized Mode. This performance improvement was particularly noticeable in non-IID data settings, as illustrated in Table \ref{tab:comparison}. An important point to highlight is that the data-sharing approach achieved a significant performance boost by sharing a mere 4-5\% of the overall data among clients. This minimal initial data sharing not only led to a significant boost in performance but also kept the communication overhead at a minimum. The personalization \& threshold filtering techniques outperformed the comparison methods in terms of precision and recall. However, they did not excel in image quality compared to the other proposed technique. Further exploration in this area as future work could unveil areas for improvement. 
    
    \subsection{Limitation}
\textbf{Sampling Time}: Diffusion models have a major drawback in terms of sampling steps and the long wall-clock time to sample synthetic data \cite{cao2022survey}. Since diffusion models are based on a Markov chain of processes that learns to noise and denoise an image through a fixed number of decay steps T (1000 or 4000), the same number of steps are required during the inference phase. This means that nearly 1000 steps are required to generate one image, which is substantially slower than GANs. Although a large number of steps are used to generate high-quality data in a controlled manner, it also comes with inherent disadvantages like increased computation complexity and longer processing times. However, several techniques~\cite{zhang2023fast,vahdat2021scorebased,xiao2022tackling} are used to speed up the notoriously slow diffusion models that could be explored for future work. 
    

   \noindent \textbf{Hardware Limitation:}
    On-device training due to FL of diffusion models provides better personalization of models and improved user privacy and security. However, large diffusion models contain over a billion parameters \cite{chen2023speed}, making it difficult to deploy in resource-constrained edge devices in FL settings. Our work employs diffusion models containing over \textbf{35 million parameters}, which is six times larger than the standard deep convolutional GANs. Some ways to overcome the increased model size without affecting performance include, model optimization~\cite{chen2023speed}, and compression~\cite{qualcomm}. 

    \subsection{Future Work}
     
    While this paper has been a start to utilizing unconditional diffusion models in a FL setting for data privacy, security, and personalization, there are several avenues for future research and exploration. One direction involves extending to real-world applications in domains like finance and healthcare,  where data is dispersed across multiple institutions and hence cannot be stored in a central server. Additionally, the development of robust evaluation metrics that go beyond just fidelity and diversity to better capture the novelty and fairness of generated samples is crucial for a comprehensive assessment of generative models. From an algorithmic standpoint, there is a need to delve deeper into privacy-preserving techniques such as differential privacy and encrypted data FL which are exciting areas of research.  Additionally, diffusion models are computationally expensive, making them difficult to deploy in edge devices as part of FL. Hence, the development of smaller models or faster inference techniques is one possible example of future work. 

    Although the results of generative AI are astonishing, many ethical concerns exist. There are concerns about intellectual property (IP) and ownership of generated content. Since generative AI models learn patterns and relationships from existing data, it is highly likely that copyrighted works are used without providing credits to the original creator. These ethical considerations can be addressed through watermarking techniques for traceability and accountability and by exploring fairness, bias, transparency, and interpretability aspects of unconditional diffusion models in the context of FL. These investigations will help shape ethical guidelines and best practices for the responsible deployment of generative models in FL settings.

\small
\input sn-article.bbl

\end{document}